\documentclass[10pt,twocolumn,letterpaper]{article}

\usepackage{cvpr}      

\usepackage{graphicx}
\usepackage{amsmath}
\usepackage{amssymb}
\usepackage{booktabs}

\usepackage[pagebackref,breaklinks,colorlinks]{hyperref}

\usepackage[capitalize]{cleveref}
\crefname{section}{Sec.}{Secs.}
\Crefname{section}{Section}{Sections}
\Crefname{table}{Table}{Tables}
\crefname{table}{Tab.}{Tabs.}

\def\cvprPaperID{*****} 
\def\confName{CVPR}
\def\confYear{2022}

\begin{document}

\title{Driver Maneuver Detection and Analysis using Time Series Segmentation and Classification}

\author{Armstrong Aboah\\
University of Missouri-Columbia\\
{\tt\small aa5mv@umsystem.edu}
\and
Yaw Adu-Gyamfi\\
University of Missouri-Columbia\\
{\tt\small adugyamfiy@missouri.edu}
\and
Senem Velipasalar Gursoy\\
Syracuse University-USA\\
{\tt\small svelipas@syr.edu}
\and
Jennifer Merickel\\
University of Nebraska Medical Center\\
{\tt\small jennifer.merickel@unmc.edu}
\and
Matt Rizzo\\
University of Nebraska Medical Center\\
\and
Anuj sharma\\
Iowa State University\\
{\tt\small anujs@iastate.edu}
}

\maketitle

\begin{abstract}
The current paper implements a methodology for automatically detecting vehicle maneuvers from vehicle telemetry data under naturalistic driving settings.  Previous approaches have treated vehicle maneuver detection as a classification problem, although both time series segmentation and classification are required since input telemetry data is continuous. Our objective is to develop an end-to-end pipeline for frame-by-frame annotation of naturalistic driving studies videos into various driving events including stop and lane keeping events, lane changes, left-right turning movements, and horizontal curve maneuvers. To address the time series segmentation problem, the study developed an Energy Maximization Algorithm (EMA) capable of extracting driving events of varying durations and frequencies from continuous signal data. To reduce overfitting and false alarm rates, heuristic algorithms were used to classify events with highly variable patterns such as stops and lane-keeping. To classify segmented driving events, four machine learning models were implemented, and their accuracy and transferability were assessed over multiple data sources. The duration of events extracted by EMA were comparable to actual events, with accuracies ranging from 59.30\% (left lane change) to 85.60\% (lane-keeping). Additionally, the overall accuracy of the 1D-convolutional neural network model was 98.99\%, followed by the Long-short-term-memory model at 97.75\%, then random forest model at 97.71\%, and the support vector machine model at 97.65\%. These model accuracies where consistent across different data sources. The study concludes that implementing a segmentation-classification pipeline significantly improves both the accuracy for driver maneuver detection and transferability of shallow and deep ML models across diverse datasets.
\end{abstract}

\section{Introduction}
\label{sec:intro}

Road crashes results in more than a million fatalities worldwide each year; on average nearly four thousand people lose their lives every day on roads (ASIRT, 2021). It is predicted that road fatalities will continue to rise to become the fifth leading cause of death in the world by 2030 (Global Status Report on Road Safety, 2021). In fact, road traffic crashes are a leading cause of death in the United States for people aged 1-54 (ASIRT, 2021). Studies have shown that about 50\% of fatal road accidents are due to unsafe driving behaviors (Global Status Report on Road Safety, 2021). Our ability to detect and characterize these unsafe behaviors in naturalistic driving settings and associate them with road accidents will be a major step toward developing effective crash countermeasures. 

Large-scale naturalistic driving studies are designed to provide insight into pre-crash causal and contributing factors. A review of the data collected from these studies can be used to extract detailed driver behavior, performance, environment information that can be associated with crashes and near crashes. NHTSA’s 100-car NDS study (100-Car Naturalistic Driving Study, 2006) was the first of many \cite{antin2019second,benmimoun2011incident,charlton2019changes,fridman2019advanced,larue2018australian} to simultaneously collect video, radar, and vehicle telemetry data from a large variety of drivers under naturalistic driving settings. To leverage this data for developing effective crash countermeasures, there is a need to annotate different types of driving events or behaviors and associate them with critical incidents, near crashes or crashes. To date, the process of driving event extraction from NDS data has been performed by a mixture of manual and semi-automated processes. The size of these datasets typically ranges between hundreds of terabytes to several petabytes depending on video compression rates. Therefore, relying on manual processing methods can be labor intensive and very expensive to scale. There is a need to develop algorithms that can ingest multi-modal NDS data and accurately annotate different driving events useful for understanding crash causality. As a result, this paper develops an end-to-end, fully automated pipeline for frame-by-frame detection and analysis of driver maneuvers from naturalistic driving videos and kinematic data. Our aim is to extract eight driving events that are critical in developing crash countermeasures: stop and lane keeping events, left-right lane changes, left-right turning movements and left-right horizontal curve maneuvers.

Existing algorithms developed for extracting driving maneuvers from NDS data can be grouped into two main categories: rule-based, pattern matching and or machine learning approaches. A rule-based algorithm is a collection of decision rules that facilitate the detection of various driving events. For instance, to distinguish between an aggressive turn and a normal turn, consider the following: if the vehicle’s heading is greater than 30 degrees, the turn is considered aggressive; otherwise, the turn is considered normal \cite{saiprasert2013detecting}. Additionally, \cite{panichpapiboon2020lane} used a rule-based algorithm to detect lane changing events. The study indicates that the lane change maneuvers will occur in three phases. The first phase is lane departure, followed by the “into” phase during which the vehicle enters the new lane, and finally the lane keeping phase during which the vehicle returns to its original position. An advantage of using this approach is that it does not require labeling of the dataset. The second class of algorithms used are pattern recognition or matching based. Algorithms are designed to extract driving events in vehicle telemetry data through a matching process against a ground-truth database of referenced driving maneuvers. The matching process is usually implemented via dynamic time warping (DTW), which compares the similarity of an incoming signal to that of a reference signal by computing a cost matrix in the form of Euclidean distance between pairwise points \cite{ali2017recognizing,panichpapiboon2020lane,saiprasert2013detecting}. The reference signal corresponding to the optimal or lowest cost path is selected as the detected driving event for the incoming or unknown signal. One of the main advantages of this technique is the ability to compare compressed and stretched portions of two signals while accounting for signal length differences.

Most recent studies \cite{bakhit2017detecting,kumar2013learning,mandalia2005using,yang2019examining,zheng2014predicting,kwakye2022social,kwakye2022travel,aboah2020smartphone} have explored the use of both shallow and deep machine learning models for driving maneuver detection and have obtain accuracies between 70 percent and 98 percent. The machine learning algorithms could either be supervised as in support vector machine (SVM), artificial neural networks (ANN), and Long-short-term-memory (LSTM) or unsupervised as in k-means \cite{bejani2019convolutional,bhoraskar2012wolverine,carvalho2017exploiting,li2021driving,yu2016fine}. Supervised learning algorithms learn and classify events based on ground truth information whereas the unsupervised learning algorithms analyze and cluster unlabeled datasets. The advantages of using ML algorithms in solving these problems are that they aid in the automation of the process of detecting driving maneuvers and produces more reliable models.

Despite significant progress and appreciable maneuver detection accuracies attained especially from ML-based algorithms, there are still open challenges that remain unsolved. First, all previous ML-based studies have treated vehicle maneuver detection as a time series classification problem. A major challenge that is not addressed by recent approaches is time series segmentation. This is an important step that should precede the development of ML classifiers: it separates any raw, continuous vehicle telemetry signal into a finite set of discrete events and anomalies with unique characteristics that can be used to train ML models for maneuver detection. The time series segmentation problem is straightforward if all the events contained in the continuous signal have a fixed duration: A simple, moving window with fixed time window could be used to define the start and end of each event. For NDS data however, the duration, frequency and amplitudes of events may vary significantly depending on the speed of the vehicle, type of sensor, driving behavior and type of event (lane change or turning movement). A robust time series segmentation algorithm is therefore needed to extract unique events that are needed to train and test ML algorithms for maneuver detection. Second, the robustness and transferability of models developed for maneuver detection have not been well tested: the size of data, number of drivers and events are usually not large enough to deduce the best performing models, or architectures needed for accurate detection of driver maneuvers. For example, \cite{mandalia2005using} and \cite{yang2019examining} reports high accuracies for only 4-431 drivers driving 8-80.4 km in the study. Studies have also evaluated these models on only one type of hardware acquisition systems: OBD or mobile phone.

As a result of the above limitations, the study develops an end-to-end pipeline for automatic, frame-by-frame labelling of NDS videos into various driving events by using vehicle telemetry data. To achieve this goal, we formulated the problem as a time series segmentation and classification problem. The segmentation task was achieved by developing a novel segmentation algorithm that utilizes the principle of energy maximization to detect the start and end of any driving event. Furthermore, the performance of both shallow and deep machine learning models for characterizing different types of drivers’ maneuvers are evaluated using a large database of NDS data (200 hours of video and vehicle telemetry data) from three different studies: SHRP2 \cite{antin2019second}, Nebraska Medical Center \cite{drincic2020digital}, and a mobile application \cite{aboah2021mobile}.  Annotating data from multiple sources enable us to evaluate the transferability of the segmentation and classification algorithms developed.

The rest of the paper is structured as follows. A review of relevant literatures is discussed in section two. Section three presents the data collection approach and problem statement. The methodology used in this study is presented in section four. Section five presents the results and discussion of the study. The study presents its conclusion and recommendation in section six.

\section{Literature Review}

This section first discusses the major naturalistic driving studies that have been conducted for the purpose of understanding driver behavior. Next, we review the different types of kinematic variables that has been used in related studies to detect and analyze driving maneuvers from vehicle on-board diagnostics (OBD) or mobile phone sensors. The last section of the review discusses different machine learning, computer vision and pattern recognition algorithms that have been developed for maneuver detection.  We explore the strengths, limitations, as well as the practical implementation of these algorithms.

\subsection{Naturalistic Driving Studies}

The 100-car NDS study is the first to collect extensive data on naturalistic driving of many drivers over an extended period. The primary goal of the study was to provide information about crashes and pre-crash events through the use of environmental and sensor data (100-Car Naturalistic Driving Study, 2006). About 100 passenger vehicles were retrofitted with a data acquisition system consisting of five cameras, a doppler radar antenna, a GPS, accelerometer, alcohol sensor, and an incident push button to continuously collect data under naturalistic driving settings. The study generated petabytes of data from 241 primary and secondary drivers, with about 43,000 hours of video data and over 3.2 million vehicle kilometers driven. The study captured many extreme cases of driving behavior including severe drowsiness, impairment, judgment error, risk taking, willingness to engage in secondary tasks, aggressive driving, and traffic violations. 

The Canadian Driving Research Initiative for Vehicular Safety in the Elderly (Candrive) conducted a similar but much larger study, with over 256 drivers, 80.5 million vehicle kilometers driven and 5 million hours of video—a total of approximately 2 petabytes of compressed data. The primary objective of the study was to identify prospectively older drivers who were medically unfit to drive \cite{charlton2019changes}. In addition to recorded videos, the study monitored participants' driving patterns by recording location data from a GPS, vehicle's speed, the position of the gas pedal, the engine's speed, and the air temperature. The study found out that elderly citizen that traveled low mileage were less prone to vehicle crashes. The European Naturalistic Driving (UDRIVE) also designed a similar study to collect data on road user behavior in various European regions under normal and near-crash conditions \cite{barnard2016study}. The study retrofitted vehicles and scooters with DAS consisting of Mobile eye smart cameras, IMU sensors, GPS, CAN data, and a sound level sensor. The type of DAS was slightly modified base on the vehicle type. For example, trucks had 8 cameras instead of 5 cameras for passenger cars. The study collected a total of 87,871 hours of video data. The Australian Naturalistic Driving Study (ANDS) aims to improve understanding of how people behave in routine and safety-critical driving situations \cite{larue2018australian}. The data for this study were gathered over a four-month period. The study recruited 360 volunteer drivers (180 from New South Wales and 180 from Victoria) and installed a data collection system in their private vehicle. The DAS is analogous to \cite{antin2019second}. The study found out that about 45 percent of the time, drivers were distracted behind the wheel.  Lastly but not the least, we discuss the MIT Advanced Vehicle Technology (MIT-AVT) which aims to set the bar for the next generation of NDS programs by leveraging large-scale computer vision analysis for human behavior \cite{fridman2019advanced}. The DAS used in this study is comprised of an IMU, GPS, and CAN messages, as well as three high-definition cameras. The research is currently ongoing and will broaden in scope in the future. 122 individuals have taken part, 15610 days have passed, 823401.5 kilometers have been traveled, and 7.1 billion video frames have been collected. The preliminary result from the study indicates that drivers tend to look at things that non-related to driving more often whiles driving.  Lastly, prior NDS emphasized vision-based approaches exclusively, omitting critical psychophysiological factors such as cognition and emotion due to technological and computing constraints \cite{tavakoli2022multimodal}. The primary objective of this study was to establish a human-centered multimodal naturalistic driving study in which driver behaviors and states are monitored using in-cabin and outside video streams, physiological signals such as driver heart rate and hand acceleration (IMU data), ambient noise, light, the vehicle's GPS location, and music logs with song features. This study is currently ongoing with no publication on the outcomes of their study.

\subsection{Kinematic Variables for Detecting Driving Maneuvers}

Recent car models have OBDs that are able to transmit high resolution vehicle kinematic information in fractions of a second. Several studies have also explored extracting and analyzing kinematic data from smartphone which tend to have a high penetration rate. Kinematic parameters such as acceleration, deceleration, orientation, yaw rate, and time to collision (TTC) are frequently used in research to identify driving events \cite{benmimoun2011incident,hankey2016description,mcgehee2007extending,lerner2010exploration,olson2009driver,pilgerstorfer2012deliverable}. The authors of \cite{benmimoun2011incident,hankey2016description,mcgehee2007extending,lerner2010exploration,olson2009driver,pilgerstorfer2012deliverable} used these parameters to examine vehicle maneuvers from NDS data. Benmimoun et al. (2011), Hankey et al. (2016), and McGehee et al. (2007) used accelerometer values to detect driving behaviors of young teens such as improper turns and curve using a rule-based approach. Olson et al. (2009) used TTC as a surrogate to measure changes in safety of the driver using rule-based approach. Pilgerstorfer et al. (2012) used lateral and longitudinal acceleration as well as TTC to assess the triggered events of truck drivers. \cite{bogard1999analysis} used GPS data to detect lane changes. \cite{xuan2006lane} proposed a similar technique for detecting lane changes by analyzing differential global positioning system (DGPS) data for the vehicle's lateral position instead. Although these are promising and much straightforward, the GPS precision levels required are not attainable from the current generation of mobile sensors. \cite{miller2005determination} in a study used yaw rate to identify lane changing maneuvers made by heavy vehicles. The researchers hypothesized that changing lanes would produce a yaw rate signal similar to that produced by a noisy sine wave. A study conducted by \cite{ayres2004method} examined both the vehicle’s velocity and yaw rate as potential variables by using a rule-based approach in detecting turns, lane changes, and curves on various types of roads. \cite{li2021driving} developed a machine learning model to detect various driving maneuvers using accelerometer and gyroscope reading using a semi-supervised machine learning algorithm.

\subsection{Approaches for Detecting Driving Maneuvers}
The approaches used in various literature for detecting driving maneuvers can be group into three categories: vision-based approaches, patten or rule-based approaches and machine learning approaches. 

All vision-based approaches begin by detecting road markings before determining any driving maneuver. The color difference between lane markings and road surfaces defines the edge, gradient, and intensity of road features used for lane detection \cite{gao2009practical,xu2009fast}. Many researchers have used edge information to find straight lines that could be lane markings in a vision-based approach to identify various driving maneuvers \cite{xu2009fast}. A B-spline is a popular mathematical model \cite{li2014lane,son2015real} which uses potential points derived from the lane markings to detect road lanes. \cite{son2015real} employs Kalman filter in tandem with a B-spline to detect lane markings. The B-spline is also commonly used to convert RGB data to HSI or custom color spaces \cite{rotaru2008color} or color features \cite{mohamed2015real}. Among the feature-based lane detection approaches are artificial neuron networks \cite{mohamed2015real}, histogram of oriented gradients (HOG) \cite{naiel2014vehicle}, and support vector machine (SVM) classifier \cite{mandalia2005using}. Although these studies produced excellent results, they do have some limitations. First, these classes of algorithms are heavily reliant on visible road markings for driver maneuver detection. Their performance suffers in the absence of lane markings, or when lane marking retroreflectivity is low. Also, vision-based approaches are affected by video quality and driving environment including weather conditions. The resolution of most NDS datasets is usually low due to high compression rates. Robust image enhancement techniques are needed to achieve modest performance for maneuver detection. 

A wide variety of machine learning approaches have been used in literature for detecting driving maneuvers. \cite{li2021driving} proposed a semi-supervised LSTM model to detect driving maneuvers. In their study, three long short-term memory (LSTM) models were built and trained to evaluate the proposed semi-supervised learning algorithm. According to the experimental results, the proposed semi-supervised LSTM could learn from unlabeled data and deliver impressive results with only a small amount of labeled data. The study compared the performance of the proposed method to other machine learning models. When compared, the proposed model outperformed existing machine learning techniques such as convolutional neural networks, XGBoost, and random forests on several measures, including accuracy, recall, F1-score, and area under the curve. The overall accuracy of the developed model was 99.7\%. \cite{bhoraskar2012wolverine} used support vector machine (SVM) to detect braking and road bumps using accelerometer, GPS, and magnetometer data collected by a smartphone and achieved an overall accuracy of 78.37\%. Júnior et al., (2017) evaluated the performance of multiple machine learning algorithms for detecting driving maneuvers (e.g., aggressive braking, acceleration, left turn, right turn) by using the area under the curve as a performance measure. The study found out that random forest outperformed other algorithms including SVM and Bayesian network with an accuracy of 99.1\%. Yu et al. (2021) employed a fully connected neural network to detect driving maneuvers such as weaving, swerving, and quick braking using the accelerometer and rotation sensors of a smartphone. The findings indicated that the neural network (95.36\%) performed more accurately than the SVM (90.34\%) at classifying driving events. \cite{bejani2019convolutional} used a convolutional neural network (CNN) to rate drivers as safe or dangerous based on accelerometer data from their smartphone. The findings demonstrated that CNN was capable of accurately classifying diverse driving styles with the use of regularization terms. The developed model achieved an overall accuracy of 95\%. \cite{carvalho2017exploiting} used recurrent neural networks to explore the detection of driving movements (RNNs). Unlike CNN, RNN was developed to learn from time series data and has shown potential. The authors compared the performance of a variety of RNN architectures, including long short-term memory (LSTM), gated recurrent unit (GRU), and standard RNN. The findings suggested that LSTM (99.7\%) and GRU (99.2\%) achieved equivalent results and outperformed traditional RNNs (93\%) in recognizing different driving maneuvers accurately.

Some studies have also used rule-based and pattern matching algorithms to detect various driving maneuvers. \cite{saiprasert2017detection} proposed both a ruled-based and pattern matching-based algorithms to detect aggressive and normal driving maneuvers. The study concludes that the pattern matching algorithm outperforms the rule-based algorithm in detecting driving maneuvers. \cite{sun2019combining}combined a dynamic time warping (DTW) and bagging tree algorithm to driving events using accelerometer and gyroscope data collected by a smartphone. Atia et al. (2017) compared the performance of K-nearest neighbor and dynamic time warping (DTW) algorithm in detecting various driving maneuvers. The results from the study indicate that k-NN achieved the best accuracy to differentiate between road anomalies and driving behaviors whereas DTW achieved the best accuracy in driving turn behavior classification.

Table 1 summarizes the literature by examining various studies, the algorithms that were used, and the kinematic variables that were used.

\begin{table*}[t]
  \setlength{\tabcolsep}{0.2\tabcolsep}
  \centering
  \small
  \caption{Literature Summary }
  \begin{tabular}{cccc}
    \toprule
    \textbf{Papers}    & \textbf{Algorithm}    & \textbf{Kinematic Variable} &\textbf{Driving Event Detection} \\
    \midrule
    \cite{miller2005determination} & FrequencyThresholding & YawRate & Lanechanges \\
    \cite{ferreira2017driver} & Rule-based algorithm  & YawRate & Lanechanges \\
    \cite{ayres2004method} & Algorithm Tunning & road’s curvature & Turns,curves\\
                                                &&&and lane change \\
    \cite{xuan2006lane} & Pattern Matching & Yaw rate, Speed & Lane changes \\    
    \cite{li2021driving} & Semi-Supervised LSTM & Differential global positioning System (DGPS) & Left-Right turns, Left-Right Lane change \\
    \cite{bhoraskar2012wolverine} & SVM & Accelerometer, Gyroscope & Break detection \\

    \cite{li2021driving} &  SVM    & Accelerometer   & Aggressive-breaking  ,right lane change \\
    &Bayesian Network&Magnetometer&left lane change,acceleration\\
    &ANN&GPS&left turn,right turn\\

    \cite{yu2016fine} & FCNN & Accelerometer, Rotation Sensors & u-turn, swerving, weaving, right turn, left turn \\
    \cite{bejani2019convolutional} & CNN & Accelerometer & Normal drive, dangerous drive \\
    \cite{carvalho2017exploiting} & RNN, LSTM, GRU & Accelerometer & Lane keeping\\
    &&&and left turn, right turn \\
    &&&and right lane change, left lane change \\
    
    \bottomrule
  \end{tabular}
  \label{tab:T1}
\end{table*}

\section{Data Collection}

Multiple streams of datasets were collected for this study. This include the Blackbox sensor data (Drincic et al. 2020), smartphone collected data (Aboah et al. 2021) and the VTTI NDS dataset (Antin et al. 2019). The Blackbox sensors developed by Digital Artefacts LLC were used to collect the data used in this study. The sensors were installed in individual personal vehicles to continuously record activities that occurred inside and outside of the vehicle. Multiple sensors, including GPS, accelerometer, wireless OBD, infrared, and high-resolution cameras, are embedded in the sensor instrumentation. As shown in Fig. 1b, the windshield-mounted sensor package, which is mounted behind the rear-view mirror. Two cameras in the system continuously capture 1) a forward view of the vehicle and 2) a view of the driver and the interior of the vehicle. The driver's behavior is continuously recorded from the time the vehicle is turned on to the time it is turned off. The study included 77 participants who were observed over a three-month period. A total of 289681.9 kilometers of data was collected across the entire United States. This dataset contains far more detailed information on driver behavior across a wide range of geographic environments than laboratory-based or retrospective studies can. The study used a developed smartphone app to collect data on both freeways and local routes. The smartphone app interface is shown in Fig. 1c. To collect data with the app, the mobile phone must first be mounted on the car's windscreen to record vehicle accelerations, rotations and some other relevant information. The video data was sampled at a frame rate of 10 frames per second, whereas the accelerometer and vehicle location data were collected at a frame rate of 30 samples per second (30 Hz). 
\begin{figure}
\includegraphics[scale=0.5]{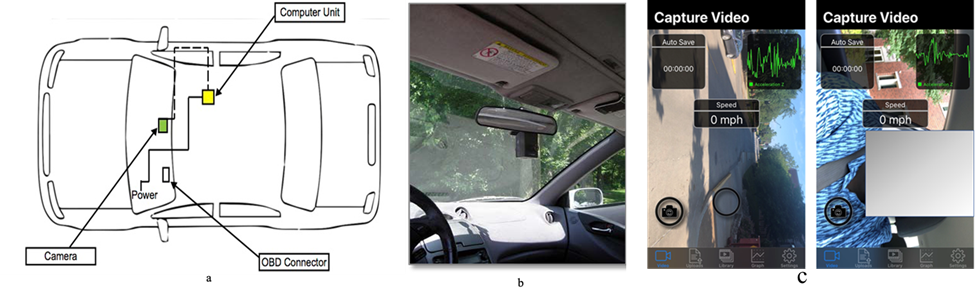}
\centering
\caption{ a) The Positioning of the Blackbox sensors in the vehicle b) Example of the Blackbox sensor inside a vehicle c) Smartphone Data Collection App Interface.}
\label{fig:p1}
\end{figure}

\subsection{Data Annotation} 
To build the benchmark dataset, the gyroscope readings were manually annotated concurrently with the driving video by noting the timestamps associated with each event in the driving video. That is, a human annotator watches the video and records the start and end timestamps of each event, after which the annotator assigns a class number to the corresponding timestamp in the signal dataset. Additionally, the study visualized both the annotated signal and the driving video concurrently to ensure that the annotations corresponded to the actual timestamp of the event. When it is determined that annotations do not correspond to actual events, they are corrected and re-visualized. This process is repeated until all annotations correspond to actual events occurring during the driving video. This method was used to obtain all ground truth labeling for the three NDS datasets that were used in this study.

\subsection{Data Cleaning and Preprocessing} 
To reduce the amount of noise in the gyroscope data, a simple moving average technique was employed to smooth the signal. As illustrated by the following (Chen, Dongyao, et al.; Karatas, Cagdas, et al.; Kang, Lei, and Suman Banerjee; You, Chuang-Wen, et al.) studies, preprocessed gyroscope signal results in gyroscope drift. The drift is responsible for shifting the actual time at which an event occurred and becomes very critical when dealing with real-time alert systems. Therefore, the study examined the occurrence of these gyroscope drifts and determined that, due to the high sampling rate of our data, the drifts were small and, as a result, did not affect our proposed algorithm's performance for extracting driving events. From Fig. 2 below, it can be seen that the gyroscope drift after preprocessing is very small and will have no effect on the study’s goal of extracting the various shapes that represent specific driving events using our proposed algorithm.
\begin{figure}
\includegraphics[scale=0.5]{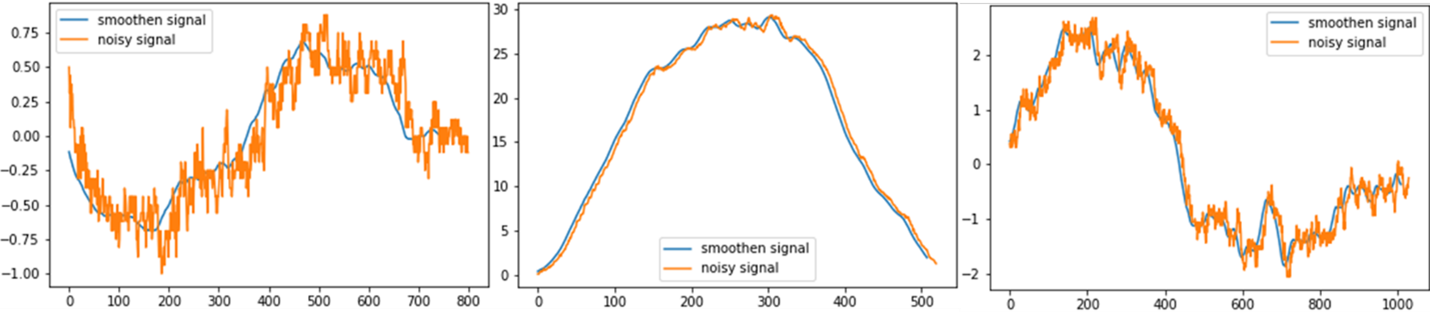}
\centering
\caption{A comparison of raw gyroscope data to processed gyroscope data}
\label{fig:p2}
\end{figure}

\subsection{Training and Validation Dataset for Developing the Classification Models} 
The study used the Nebraska NDS dataset to develop all four models. All four models were trained on an NVIDIA GTX 1080ti GPU with 16,233 training samples. For all developed models, we used a 70:30 split for model training and validation. The training samples are distributed as follows; right turns- 4,362 samples, left turns- 3,968 samples, right curves- 2,051 samples, left curves- 1,947 samples, right lane change- 1,895 samples and left lane change- 2,010 samples.

\section{Methodology} 
\subsection{Problem Formulation and Overview} 

Most existing methods for driver maneuver detection formulates the problem purely as a classification problem, assuming a discretized input signal with known start and end locations for each event or segment. However, in practice, vehicle telemetry data used for detecting driver maneuvers are continuous, therefore, a fully automated driver maneuver detection system should implement solutions for both time series segmentation and classification. The method proposed in this paper maps a continuous sequence into a dense segmentation followed by event classification using machine learning and a heuristic algorithm. 

Specifically, let $x\in R^{\tau S\times C} $ represent a vehicle telemetry dataset with C sensors or channels, sampling at a rate S for a period of $\tau$ minutes. Let ${\mathbf{v}}=v_1,v_2,v_3,\ \ldots,\ v_n\} $ be a video sequence of frames that corresponds to each sample in x.  The goal is to map x into finite set of segments, $\left\lfloor\tau\bullet e\right\rfloor\ $ where e is the segmentation frequency. Compared to other segmentation approaches (Perslev et al., 2019.) where e is fixed, in the current method, the parameter is variable and adaptively selected based on input signal features. Each segment is passed through a model $f\left(\tau\bullet e;\theta\right):\mathbb{R}^{T\times i\times C}\rightarrow\mathbb{R}^{T\times K} $ with parameters $\theta$ that maps each segment $\left\lfloor\tau\bullet e\right\rfloor $ to one of K class labels including: stop and lane keeping events, lane changes, left-right turning movements and horizontal curve maneuvers. Finally, frame-by-frame annotation of video sequence is achieved by mapping classified segments to the image domain.  

The general methodology adopted for automatic detection of driver maneuvers consists of 5 distinct steps as shown in Fig. 3. First, multi-modal data is pre-processed and standardized, followed by segmentation of kinematic data into main driving events, anomalies and lane keeping events. Anomalies and lane keeping events are passed through a heuristic’s algorithm which further classifies these events into anomalies, stop and lane-keeping events. Only driving events are passed through a machine learning classifier. By training only main driving events, the ML classifiers were able to learn the unique characteristics of lane changing and turning movement events without confusing them with other features such as lane-keeping, lane-incursions events and anomalies caused by road roughness or erratic driving behaviors. The outputs of the classifiers and heuristics are finally used for frame by frame driving event annotation of raw video feeds. Each step of the methodology is further discussed in the sections below.

\begin{figure}
\includegraphics[scale=0.5]{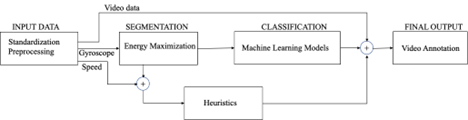}
\centering
\caption{Flowchart of Methodology}
\label{fig:p3}
\end{figure}

\subsection{Input Data Normalization} 
Although previous studies used kinematic variables such as yaw rate, accelerometer readings to detect limited driving maneuvers, the current study determined that gyroscope reading (which measures the orientation and angular velocity of the vehicle), and the vehicle's speed data are the two key input variables that can be used to characterize all driving maneuvers consistently across different hardware measurement systems. The gyroscope readings (z-axis) were smoothed using a simple moving average and subsequently feature-scaled with the mean normalization equation defined in Equation 1. The raw speed data will be used to develop heuristics for detecting lane keeping and stopped events whereas the standardized gyroscope readings are pushed through a time series segmentation algorithm for driving maneuver event detection.

\begin{equation}
    \hat{x}=\frac{x_i-\bar{x}}{x_{\max }-x_{\min }}
  \label{eq:eq1}
\end{equation}

Where $x_i$ is the gyroscope reading at timestamp $i$ , $\bar{x}$ is the mean of all data, $x_{max}$ is the maximum data value and $x_{min}$ is the minimum data value. 

\subsection{Time Series Segmentation } 
The segmentation step involves the extraction of driving events using the energy maximization algorithm (EMA).  The fundamental assumption driving EMA is that the sum of the energy from the start of an event will continuously increase until the end of the event is reached. At each time step, t , we dilate a moving window at different rates of w. For each dilation rate, $w_i$, the energy of $X\left[t-\frac{w_i}{2}:t+\frac{w_i}{2}\right]$, is computed using Equation 2. In Equation 2, the computed energies are scaled by a factor of  $\frac{S}{N}$. This factor takes into account the duration of the event (S) and the number of data points (N) in the signal so that events that are not fully captured but has a greater energy could be penalized. 

\begin{equation}
    e_n=\frac{s}{N} \sum_{n=0}^{n+1} X[n]^2
  \label{eq:eq2}
\end{equation}

We then determine if an event is present based on the calculated energies. The dilation rate, w, increases by a factor of 0.25s and continues to dilate until the computed energy is a maxima, i.e., $E_{t-n}\le\ E_t\geq\ E_{t+n}$.  If more than one maximum is detected, we use non-maximum suppression to remove overlapping signals. The whole process is repeated for each time step until the final time step for the input signal. The output from the segmentation step is either an event or non-event as shown in Fig. 4.

\begin{figure}
\includegraphics[scale=0.5]{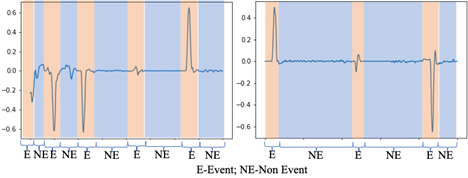}
\centering
\caption{Segmentation of Signal into Events and Non-events }
\label{fig:p4}
\end{figure}

Fig. 5. shows an example of a time step t  and a plot of energies computed at different dilation rate, w. The step in performing the segmentation is summarized in Table 2 below. Events detected through the energy-maximization process are events are passed through the ML models for classification into lane changes and turning movements. Non-events are passed through a heuristics algorithm for classifying lane-keeping and stop events.

\begin{figure}
\includegraphics[scale=0.5]{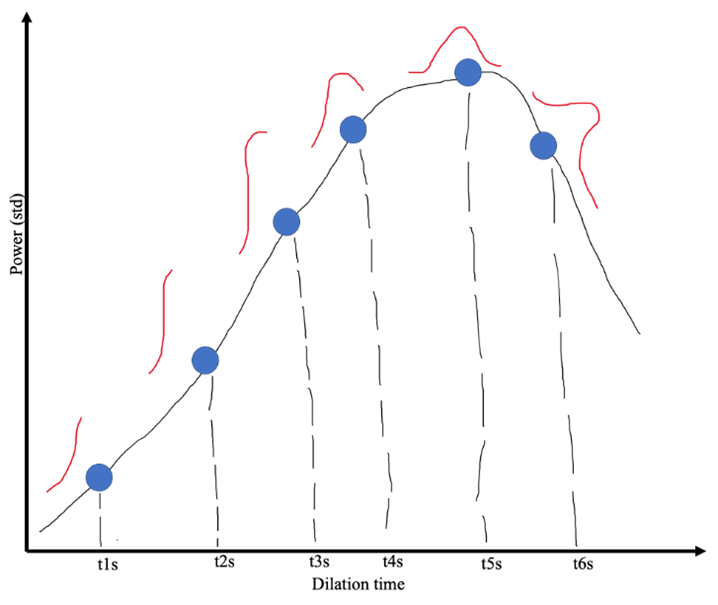}
\centering
\caption{An illustration of the dilating time window }
\label{fig:p5}
\end{figure}

\begin{figure}
\caption*{Table 2: Pseudo code for Performing Event Segmentation}
\includegraphics[width=0.5\textwidth,scale=0.5,height=6cm]{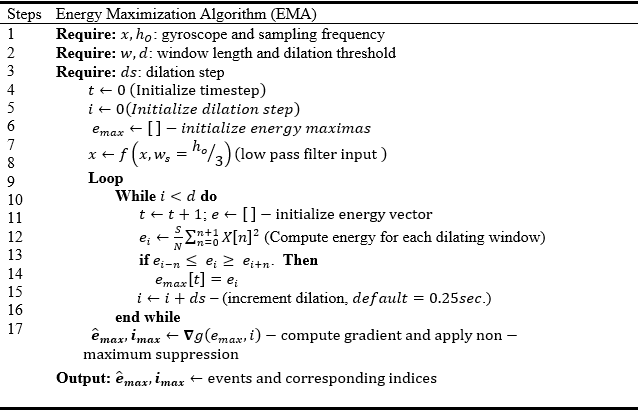}
\centering
\label{fig:p15}
\end{figure}

\subsection{Classification } 
Supervised ML algorithms were used to classify the events extracted from the EMA into six maneuvers: left-right lane changes, left-right turning movements and left-right horizontal curves movements. Four main classifiers were evaluated including: LSTM, SVM, 1D-CNN, Random Forest. The model architectures, and structure of the input data used for training each machine learning algorithm are explained in detail below.

\subsubsection{Input Data Structure}
The extracted events from the segmentation step are restructured before being passed to the classification algorithms. For the deep learning algorithms, the raw samples from the energy maximization algorithm are passed through them for classification, whereas for the machine learning algorithms, the raw data is first passed through a principal component analysis (PCA) to reduce dimensionality before being passed through them. The number of input features and output features are 50 and 1 respectively for all trained models. A summary of the training parameters used to build the various models are presented in Table 3 below.

\begin{table*}[t]
  \setlength{\tabcolsep}{1\tabcolsep}
  \centering
  \small
  \caption*{Table 3: Training Parameters}
  \begin{tabular}{cccc}
    \toprule
    \textbf{Parameters}    & \textbf{LSTM}    & \textbf{SVM} &\textbf{1D-CNN} \\
    \midrule
    Number of features & 1 & 1 & 1 \\
    RNN Layers & 3  & X & X \\
    Hidden Layers & 20 & X & 250\\
    Learning rate & 0.001, X & 0.001 \\    
    Loss function & Cross entropy loss & X & Cross entropy loss \\
    Optimizer & Adam & X, Adam \\
    Activation function &  X    & X   & ReLU and Sigmoid\\
    Epoch & 600 & X & 40 \\
    Kernel & X & Linear & X \\
    Gamma & X & Auto & X\\
    Kernel size & X & X & 3 \\
    Filters & X & X & 250 \\
    \bottomrule
  \end{tabular}
  \label{tab:t3}
\end{table*}

\subsubsection{Long-short-term-memory (LSTM) model}
LSTM is a supervised deep learning architecture which is used for both classification and regression. LSTM is a special type of recurrent neural network (RNN) that is capable of learning long-term dependencies (Li et al. 2021). It has a single cell state that runs the length of the chain. The cell state can be modified by either adding or removing information. The LSTM architecture comprises of three gates that protect and control information that pass through the cell state. These gates are the forget gate, the input gate, and the output gate. 

The forget gate deletes information from the cell state that is not required to pass through to the input gate. This is accomplished by the equation below.

\begin{equation}
    f_t=\sigma\left(W_f \cdot\left[h_{t-1}, x_t\right]+b_f\right)
  \label{eq:eq3}
\end{equation}

At the input gate, new information is stored, and values are updated in the cell state. This results in the creation of a vector of new candidate values, ${\widetilde{C}}_t$. The mathematical representation of what happens in the input gate is shown in the equations below.

\begin{equation}
    \tilde{C}_t=\tanh \left(W_c \cdot\left[h_{t-1}, x_t\right]+b_c\right)
  \label{eq:eq4}
\end{equation}

The cell state is updated by adding the previously deleted information,$\ f_t*C_{t-1} $ to the newly added information $i_t*{\widetilde{C}}_t $. The updated cell state can be expressed mathematically as 

\begin{equation}
    C_t=f_t * C_{t-1}+i_t * \tilde{C}_t
  \label{eq5}
\end{equation}

Finally, the output gate outputs the relevant portions of the cell state.

\begin{equation}
    o_t=\sigma\left(W_0 \cdot\left[h_{t-1}, x_t\right]+b_0\right)
    h_t=o_t * \tanh \left(C_t\right)
  \label{eq:eq6}
\end{equation}

\subsubsection{Support Vector Machine (SVM)}
Support Vector Machines (SVMs) are a type of supervised machine learning technique that can be used to solve both regression and classification problems. SVMs are designed to strike a balance between fitting the training data and reducing model complexity (Cortes et al. 1995). This method of defining a loss function is known as structural risk minimization (SRM), and it typically yields a better model generalization than the empirical risk minimization approach of defining a loss function. SVMs were originally developed to solve two-group classification problems; therefore, applying it to multi-label classification problems results in the input data being highly dimensional.  As a result of the high dimensionality of the input data, computational issues such as handling large vectors and overfitting occur. These issues are resolved by the addition of a kernel function. A kernel function returns the dot product of the original data points' feature space mappings. SVMs employ a variety of kernel functions, including linear, polynomial, and Gaussian RBF. The algorithm for performing multiclass classification using support vector machines involves transforming the input vector into a higher-dimensional feature space. In the feature space, a linear decision surface called a hyperplane is constructed (Cortes et al. 1995). The hyperplane represents the greatest separation between any two classes. In addition, two parallel hyperplanes are constructed on either side of the hyperplane to segregate the data. A separating hyperplane is one that minimizes the distance between two parallel hyperplanes as shown in Equation 7.

\begin{equation}
    \max _v \frac{1}{2}\left(\min _{x_i \in C_1} v^T\left(x_i-x_0\right)-\min _{x_j \in C_2}(-v)^T\left(x_j-x_0\right)\right)
\end{equation}
\begin{equation}
   \text { s.t }\|v\|_2=1
\end{equation}
Where v is the unit vector, $C_{1\ }$and$\ C_2$ are contants, and  $x_{i\ }\in\ R^k$. The idea is that the larger the margin or distance between these parallel hyperplanes, the smaller the generalization error of the classifier.

\subsubsection{1D-CNN}
A one-dimensional convolutional neural network (1D-CNN) is a deep learning architecture that can either be supervised or unsupervised and can be used for both regression and classification problems. The A 1D-CNN applies a kernel along a one-dimension input data. The input data is usually a signal data with two dimensions.  The first dimension is the time-steps whereas the second dimension is the signal values. Fig. 6 shows the flowchart of the 1D-CNN. Mathematically, a one-dimensional convolutional neural network is composed of an input vector $x\ \in\ R^p$ and a filter $w\ \in\ R^k$ where $k\ \le p$. The 1D-CNN takes $w^Tx\left[i:i+k\right]$ for each surrounding set of k elements of $xx\left[i:i+k\right]$ and gives one node of the convolutional layer.

\begin{figure}
\includegraphics[scale=0.5]{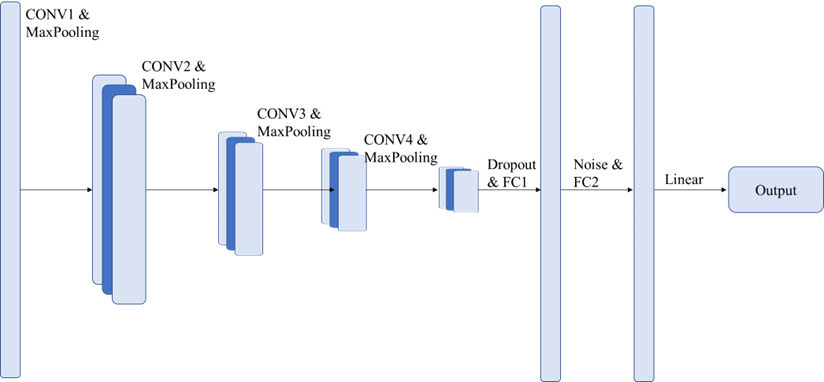}
\centering
\caption{1D-CNN Architecture }
\label{fig:p6}
\end{figure}

\subsubsection{Random Forest}
A random forest is a supervised machine learning algorithm that leverages ensemble learning to solve complex problems involving regression or classification. Random forests are a collection of tree predictors in which the values of a random vector are sampled independently and uniformly across the forest to determine the values of each tree (Breiman 2001).  As the number of trees in a forest increase, the generalization error converges to a limit (Breiman 2001). The generalization error of a forest of tree classifiers is proportional to the strength of the trees in the forest and their correlation. By randomly splitting each node on a tree, we obtain error rates that are comparable to Adaboost's (Freund et al. 1996), but more robust to noise in this case (Breiman 2001). Internal estimates of error, strength, and correlation are used to demonstrate the effect of increasing the number of features used in the splitting process. Internal estimates are also used to determine the significance of variables. Each class in the dataset is determined by letting all the trees in the forest vote for a class. The most voted class becomes the classification of the data points.

For each internal nodes of the tree, it takes a subset of features at random and utilizes that information to compute the centers of various classes present in the data at the current node. For example, given two classes, 0 and 1, the centers of the classes will be denoted as Left-Center and Right-Center respectively.

\begin{equation}
left_{center}\left[k\right]=\ \frac{1}{n\ }\sum_{i=1}^{n}{x_{ik}I\left(y=0\right)}
\end{equation}
\begin{equation}
right_{center}\left[k\right]=\ \frac{1}{n\ }\sum_{i=1}^{n}{x_{ik}I\left(y=1\right)}
\end{equation}

Where $I\left(y=0\right)$ and $I\left(y=1\right)$ are the dictator functions. Each record in the dataset is assigned to the appropriate class at the present node by computing the Manhattan distance between the center and the record as illustrated in Equation 10.

\begin{equation}
Distance\ \left(center,\ record\right)=\ \sum_{i\in s\ u\ b}\left|ceter\left[i\right]-record\left[i\right]\right|            
\end{equation}

Where sub is the subset attributes randomly selected from the dataset. Each tree, therefore, grows without pruning.

\subsubsection{Heuristics}
The characteristic patterns of vehicle telemetry data especially during stop and lane keeping events vary widely even for the same driver. As a result, it generates high false positive rates when fed through machine learning models. In the current study we developed a heuristic algorithm based on the vehicle speed and an adaptive thresholding technique to classifying lane-keeping and stop events. A stopped event occurs when the speed of the vehicle is zero. To detect lane-keeping events, we draw from a probability distribution curve. The assumption here is that lane-keeping is the most dominant event in every trip. Therefore, all gyroscope readings about k standard deviations from the mean should belong to this class. A value of k=2 was used in this study. The equation below summarizes the heuristic algorithm. 

\begin{equation}
e_l=\left\{\begin{matrix}stop\ \ \ \ \ \ \ \ \ \ \ \ \ \ if\ speed\cong0\\lane-keeping\ \ \ \ \ \mu+k\sigma\le x\le\mu+k\sigma\ \ \ \ \ \ \ \ \ \ \ \ \\\end{matrix}\right.           
\end{equation}

\subsection{Video Annotation}
The goal of this methodology is to automate the frame-by-frame annotation of driving events of NDS dataset. The machine learning classification outputs the start and end time of the event which is the same as the heuristics. The classification outputs and indices from the machine learning models and heuristics are combined and transferred into the time domain for frame-by-frame video annotation.

\section{Results}
\subsection{Performance Measures}
The efficiency and accuracy of the Energy Maximization Algorithm and the various machine learning models were evaluated using various performance measures. We assessed the performance of the machine learning models using precision (P), F1 score (F1), and recall value (R). The F-1 score is the harmonic average of the recall and precision values. Precision is defined as the ratio of true positives (tp) to all predicted positives (tp+fp), as shown in Equation 13. Similarly, recall is the ratio of true positives to all true positives (tp+fn) is defined in Equation 14.

\begin{equation}
 Precision=\ \frac{tp}{tp+fp}                      
\end{equation}
\begin{equation} 
 Recall=\ \frac{tp}{tp+fn}                     
\end{equation}
\begin{equation}
 F1=\ \frac{2*Precision*Recall}{Precision+Recall}                       
\end{equation}

On the other hand, we evaluated the accuracy of the EMA using a duration score (DS) metrics computed as

\begin{equation}
e_l=\left\{\begin{matrix}1\ \ \ \ \ \ \ \ \ \ \ \ \ \ \ \ \ if\ \left|A_i-P_i\right|>1\\\left|\ A_i-P_i\right|\ \ \ \ \ \ \ \ \ else\ \ \ \left|A_i-P_i\right|<1\ \ \ \ \ \ \ \ \ \ \ \ \\\end{matrix}\right.                    
\end{equation}

\begin{equation}
DS=1-\frac{1}{n}\sum_{i=1}^{n}e_l                      
\end{equation}

Where $A_i$ is the ith actual event duration, $P_i$ is the ith extracted event duration by the EMA, and n is the number of total events. Finally, the overall accuracy of the pipeline was calculated using Equation 16,17 and 18. The F1 scores are multiplied by the duration score in this performance metric.

\begin{equation}
Overall\ Accuracy=F1\ score*DS                     
\end{equation}

\subsection{Segmentation Outputs}
The segmentation step outputs either an event (turns, lane changes, and curves) or a non-event (lane keeping and stop) as show in Fig. 4. The EMA in the segmentation step extract very distinctive shapes from the signal as shown in Fig. 7. From Fig. 7., it can be observed that the full extent of signals corresponding to turns, lane changes, curve negotiation, are extracted at different length of time. Also, all the various driving events have varying amplitudes as shown in the Fig. 7. Events that occur in both directions (right and left) have similar shape and amplitude but occur at different phases. Example is the right turn and left turn. The algorithm’s ability to extract the full shape of the signal before taken into the model alleviate the limitation encountered with the fixed time window approach leading to a lot of false positives. Additionally, lane changes are clearly differentiated from lane keeping due to their distinctive shapes. The extracted events are classified using machine learning models, while the non-events are classified using heuristics.

\begin{figure}
\includegraphics[scale=0.5]{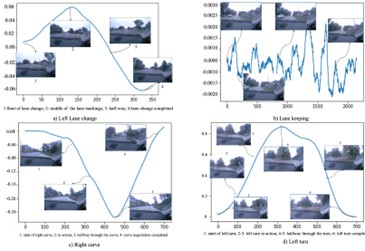}
\centering
\caption{Extracted Events Shape by the EMA}
\label{fig:p7}
\end{figure}

\subsection{Evaluating the Accuracy of The Energy Maximization Algorithm}
The actual durations of events were compared to the EMA-derived durations. According to Fig. 8., the distribution of event durations for actual and extracted events was similar for right turns, right curves, and left curves. Additionally, some extracted left turns had durations that were significantly longer than the actual left turn durations. This can be explained by the fact that some left curve or right curve negotiations are immediately followed by a left turn in which the EMA records a portion of those events as left turns, resulting in the increased duration of some left turns. In general, it is observed that the durations of events extracted via EMA are significantly longer than the durations of events manually extracted.

\begin{figure}
\includegraphics[scale=0.5]{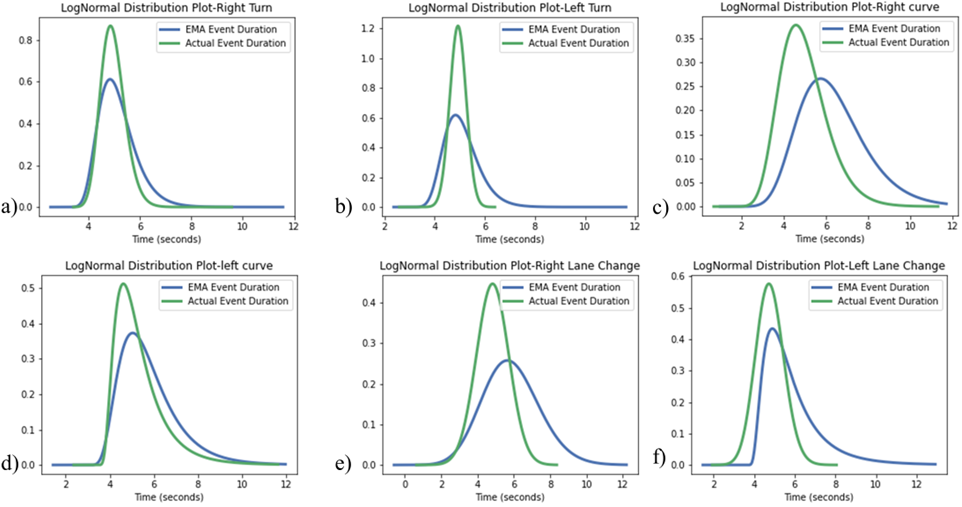}
\centering
\caption{Comparative Analysis of Lognormal Distribution of Event Duration a) right-turns b) left-turn c) right-curves d) left-curve e) right-lane-change f) left-lane-change}
\label{fig:p8}
\end{figure}

Additionally, the extracted time distributions were compared to time distributions extracted for a variety of events in previous research. According to a study by (Toledo et al. 1999), lane change durations range between 3 and approximately 7 seconds, which is consistent with the results shown in Fig. 9e and Fig. 9f. Additionally, the majority of right and left turns occur within the range of 4-6 seconds. For right curves, the majority of durations fell within the range of 4-6 seconds, but a sizable portion fell within the range of 7-10 seconds. These variations are explained by the varying lengths of right curves observed at various locations. Certain right curves are longer than others, requiring vehicles to negotiate for a longer period of time. For left curves, the same is true. Additionally, the majority of lane changes occurred within the range of 3-5 seconds, which is consistent with the findings of (Toledo et al. 1999).

\begin{figure}
\includegraphics[width=0.4\textwidth]{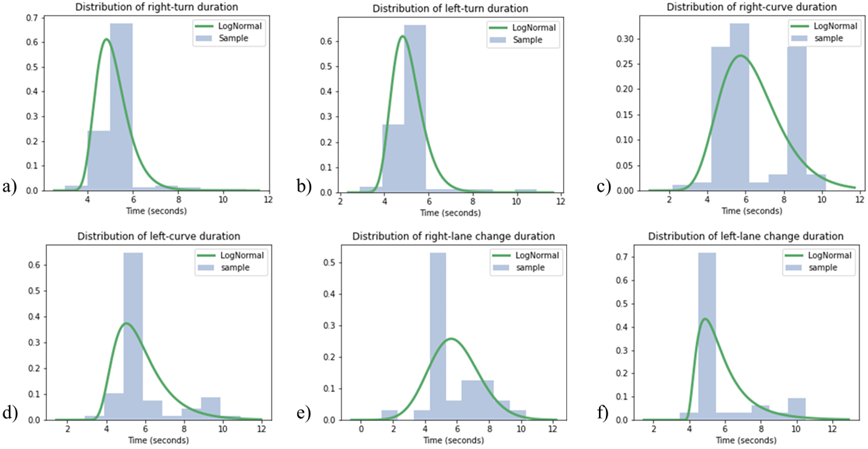}
\centering
\caption{Lognormal Distribution of Event Duration a) right-turns b) left-turn c) right-curves d) left-curve e) right-lane-change f) left-lane-change}
\label{fig:p9}
\end{figure}

Finally, using Equation 15, the accuracy of the EMA was calculated, and the results are summarized in Table 4. According to Table 4, right and left turns, as well as right and left changes, had significantly high accuracies. On the other hand, the accuracies of the left and right curves were relatively low. The low accuracies can be attributed to a variety of factors, including the algorithm treating two consecutive events as one, as it is typically observed when a left curve follows a right curve or vice versa.

\begin{table}
  \setlength{\tabcolsep}{1\tabcolsep}
  \centering
  \small
  \caption*{Table 4: Accuracy of the EMA based on Durations of Extracted Events}
  \begin{tabular}{cc}
    \toprule
    \textbf{Driving Maneuvers}    & \textbf{Accuracy of EMA} \\
    \midrule
    Right turn & 0.820 \\
    Left turn & 0.843  \\
    Right curve & 0.654 \\
    Left curve & 0.690 \\    
    Right lane change & 0.618 \\
    Left lane change &  0.593 \\
    Lane-keeping & 0.856 \\
    Stop & 0.848 \\
    \bottomrule
  \end{tabular}
  \label{tab:t4}
\end{table}

\subsection{Classification Results}
\subsubsection{Model Comparison}
In this study, four machine learning models were developed. Our analysis revealed that all four models had accuracies comparable to those reported in studies that trained similar models using a variety of kinematic variables (Bakhit et al. 2017; Kumar et al. 2013; Mandalia and Salvucci 2005; Zheng et al. 2014). It can be deduced that the gyroscope reading is sufficiently sensitive to detect all driving events, as seen when other kinematic variables are combined to perform the same task (Bhoraskar et al. 2012). Additionally, when comparing the number of iterations required to train the deep learning models, the 1D-CNN model converges after 20 epochs, whereas the LSTM model converges after 300 epochs. The 1D-CNN model, therefore, trains faster than the LSTM model. When the accuracies of all four models were compared, the overall accuracy of the 1D-CNN model was 98.99 percent, followed by the LSTM model at 97.75 percent, then RF model at 97.71 percent, and the SVM model at 97.65 percent that are comparable to accuracies obtained by (Bakhit et al. 2017; Kumar et al. 2013; Mandalia and Salvucci 2005; Zheng et al. 2014). The consistency of the accuracies obtained for all four models indicates that the EMA is effective at capturing all driving events. Furthermore, we evaluated the performance of all models using the F1 score, precision, and recall values for each driving maneuver as shown in Table 5. Lane change maneuvers (both left and right) had low F1 scores across all models. This is because of the false negatives caused by missed lane change events, which are particularly prevalent on highways with relatively high speeds. Additionally, right turn maneuvers had the highest average F1 scores across all models, ranging from 0.991 to 0.998. Similar scores were observed for left turns and right-left curve negotiations. Lane keeping on the other hand, had a high rate of false positives due to missed lane changes, particularly on highways and also stops. In summary, all models performed similarly well at predicting all types of driving maneuvers, with fewer false positives and negatives. Table 5 summarizes the models' predictions for specific driving events.

\begin{figure}
\caption*{Table 5: Precision, Recall and F1 Values Obtained From LSTM, SVM, 1D-CNN, and RF}
\includegraphics[scale=0.5]{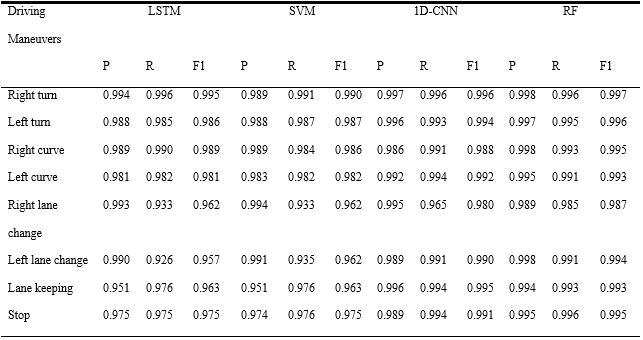}
\centering
\label{fig:p10}
\end{figure}

The study further examined the overall accuracy of the developed pipeline using Equation 16, and the results are summarized in Table 6. In this performance metric, the F1 scores were penalized by the duration scores. Overall accuracy per driving event ranges between 0.645 and 0.852, as shown in Table 6. Right and left curves both exhibits relatively low overall accuracy, owing to their low duration score values.

\begin{figure}
\caption*{Table 6: Overall Accuracy of Pipeline}
\includegraphics[scale=0.5]{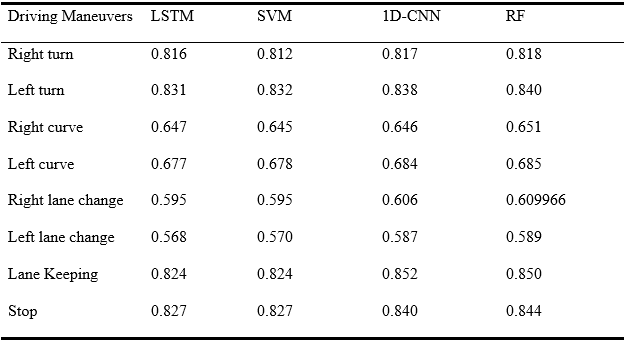}
\centering
\label{fig:p11}
\end{figure}

\subsubsection{Comparative Analysis: Proposed Methodology vs Fixed Time Window Approach}
To further investigate the effectiveness of the EMA in extracting driving events and its relevance in the proposed methodology, the study compared the detection outcome of an EMA-extracted event to the detection outcome of a fixed time moving window approach on a continuous signal, which has been used in several studies (Houenou et al. 2013; Morris et al. 2011; Ohn-Bar et al. 2014). We considered two different fixed time window methods: the three-second moving time window approach and the five-second moving time window approach. The results indicate that the energy maximization algorithm produced consistent results across all three models, whereas the fixed time window approach did not. Also, as shown in Table 7, the 5-second fixed moving time window performed better than the 3-second moving time window, which is consistent with results in studies (Houenou et al. 2013; Morris et al. 2011; Ohn-Bar et al. 2014). On a continuous signal, the proposed methodology outperforms fixed moving time window approaches for detecting driving events. 

\begin{figure}
\caption*{Table 7: Overall Test Accuracy}
\includegraphics[scale=0.5]{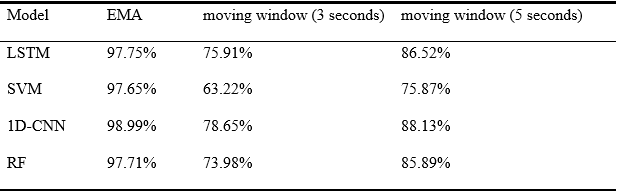}
\centering
\label{fig:p12}
\end{figure}

\subsection{Test of Model’s Transferability }
To test the transferability of the models developed, specifically the 1D-CNN, the study evaluated the developed model on events extracted from three different data sources, namely the SHRP2 NDS dataset (Antin et al. 2019), the Nebraska Medical Center NDS dataset, and data collected via a smartphone (Aboah et al. 2021). The study analyzed about 150 video hours of SHRP 2 dataset, 200 hours of Nebraska Medical Center NDS dataset and 100 hours of smartphone collected dataset. Table 8 summarizes the outcomes of the predictions for all three datasets. The F1 scores were high and consistent across all three datasets. The results implies that the developed model and algorithm are easily transferable to predict driving event from signal data from different sensor types.

\begin{figure}
\caption*{Table 8: Comparison of Precision, Recall and F1 Values Obtained All Three Datasets}
\includegraphics[scale=0.5]{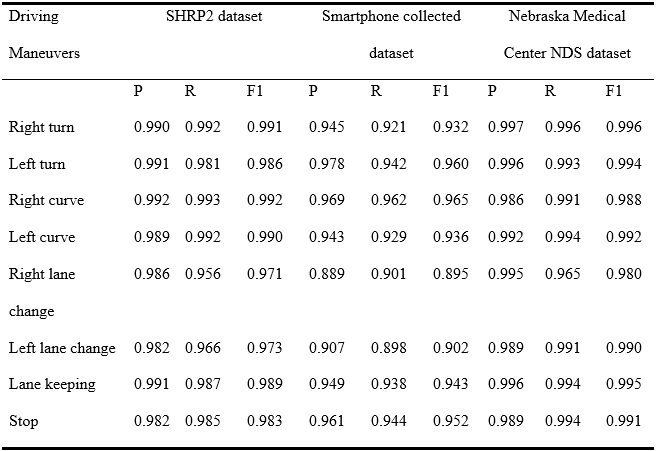}
\centering
\label{fig:p13}
\end{figure}

Additionally, all four models were combined to create a decision tree-like structure. Where each branch of the tree is a representation of a different model. Extracted events from the segmentation step are classified by passing them through each branch. Following that, the branches vote on the most frequent class. The Table below summarizes the analysis's findings. As seen in the Table 9, there is a slight improvement in both precision and recall values for all classes on average compared to relying on a single model prediction as illustrated in Table 8 for all three datasets.

\begin{figure}
\caption*{Table 9: Combined Model Performance All Three Datasets}
\includegraphics[scale=0.5]{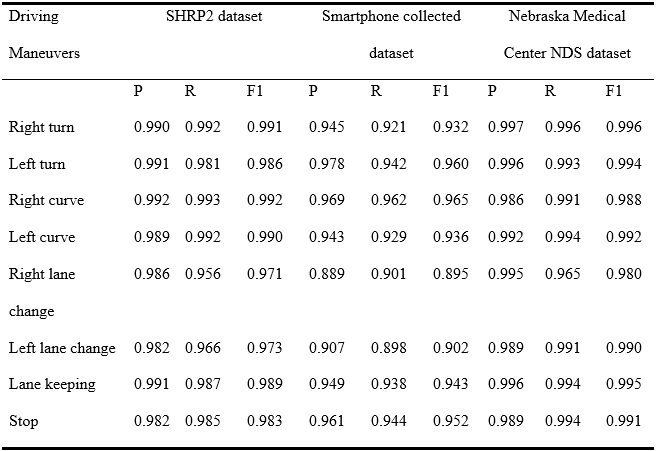}
\centering
\label{fig:p14}
\end{figure}

\subsection{End-to-End Pipeline for Annotating NDS Videos}
Finally, the study developed an end-to-end pipeline that takes the NDS video, gyroscope reading, and vehicle speed as inputs and outputs an annotated video of driving events, as illustrated in [https://youtu.be/JAuCfRGnLBI]. To annotate each video frame, the indices of the time series (segmented and classified) are aligned with the video stream, taking into account differences in sampling rates. We interpolate and upsample the vehicle telemetry data if its sampling frequency is higher than the videos frame rate and vise-versa.

\subsection{Application of Research Findings}
The primary application of this research is to develop crash countermeasures by better understanding drivers’ behaviors in naturalistic settings, specifically, the drivers’ environment. The results from this study, when conducted on large-scale, will provide insight and the extraction of some critical information such as drivers' lane-changing behaviors (which have recently been the cause of the majority of vehicle crashes on the highway) and turning maneuvers, as well as aggressive driving behaviors, for the purpose of improving traffic safety. The framework for this study is both fast and scalable. As such the framework developed in this study is going to facilitate the annotations of large-scale of NDS videos into various driving events. The extraction of these events will allow for more rapid analysis of conflict zone crashes especially at intersections (i.e., the extraction of right and left turning events). 

\section{Conclusion}
To effectively use NDS data to deduce crash causation, algorithms must be developed that can ingest multi-modal NDS data and annotate various driving events pertinent to deducing crash causation. Recent studies have examined the use of shallow and deep machine learning models for driving maneuver detection, obtaining accuracies ranging from 70\% to 98\%. A significant limitation that these ML approaches do not address is the time series segmentation problem. The current study addressed this limitation by 1) developing an energy maximization algorithm (EMA) that is capable of extracting distinct shapes of driving events from telemetry data. Also, the effectiveness of the EMA was further investigated through the development of four machine learning models. 

Multiple sources of data were used in this study including Blackbox sensor data, smartphone data and VTTI dataset. The study accomplished its objectives through the development of a five-stage methodology: 1) preprocessing of data, 2) event segmentation, 3) machine learning classification, 4) heuristics classification, and 5) frame-by-frame annotation of video. To begin, the input data is standardized and smoothed. The resulting output is segmented and then classified using both machine learning (main driving events) and heuristics (stops and lane-keeping). The study separated the detection of stops and lane-keeping from the rest of the driving events because the two can be easily identified using simple thresholding and to reduce false negatives when using only ML to classify all driving events.

The result from the study indicates that the gyroscope reading is a very good parameter to be use in extracting driving events since it showed consistent accuracy across all four developed models. The study shows that the accuracy of the Energy Maximization Algorithm ranges from 56.80\% (left lane change) to 85.20\% (lane-keeping) All four models developed had comparable accuracies to studies that used similar models (Bakhit et al. 2017; Kumar et al. 2013; Mandalia and Salvucci 2005; Zheng et al. 2014). The 1D-CNN model had the highest accuracy of 98.99\%, followed by the LSTM model at 97.75\%, the RF model at 97.71\%, and the SVM model at 97.65\%. To serve as a ground truth, continuous signal data was annotated. Also, the proposed methodology outperformed the fixed time window approach when compared. The study further analyzed the accuracy of the overall pipeline by penalizing the F1 scores of the ML models with the duration score of the EMA. The overall accuracy of the pipeline was in the range of 56.8\% to 85.2\%. To test the model's transferability, the developed models were used to detect driving events from multiple streams of datasets. The F1 scores were high and consistent across all three datasets used. The predicted results were compared to the ground truth annotations. Using the LSTM model, the test was 91\% accurate.
The study did not take advantage of large database of video data acquired; Future work should consider integrating the video data, with other predictive models such as eye detection model, and object detection models to better understand the driver’s behavior.

\subsection{Limitations to Study}
One of the challenges encountered in this study was dealing with outliers due to anomalous behaviors of drivers. The data outliers are due to false spikes in the gyroscope readings caused by the driver's activity in the vehicle. For instance, a spike in the gyroscope reading can be observed when the driver is dancing or drinking while keeping a lane or at a stop. While these spikes are not considered events, the EMA will extract them as events and pass them through the classification algorithm. These outliers contribute to the increased detection of false positives.

{\small
\bibliographystyle{ieee_fullname}
\bibliography{egbib}
}

\end{document}